\documentclass[]{spie}  

\usepackage{amsmath,amsfonts,amssymb}
\usepackage{graphicx}
\usepackage[colorlinks=true, allcolors=blue]{hyperref}

\title{Visualization Framework for Colonoscopy Videos}

\author[]{Saad Nadeem}
\author[]{Arie Kaufman}
\affil[]{Department of Computer Science, Stony Brook University, Stony Brook, NY, 11794, USA}

\authorinfo{Further author information: Send correspondence to Saad Nadeem. E-mail: sanadeem@cs.stonybrook.edu}

\begin{document}
\maketitle

\begin{abstract}
We present a visualization framework for annotating and comparing colonoscopy videos, where these annotations can then be used for semi-automatic report generation at the end of the procedure. Currently, there are approximately 14 million colonoscopies performed every year in the US. In this work, we create a visualization tool to deal with the deluge of colonoscopy videos in a more effective way. We present an interactive visualization framework for the annotation and tagging of colonoscopy videos in an easy and intuitive way. These annotations and tags can later be used for report generation for electronic medical records and for comparison at an individual as well as group level. We also present important use cases and medical expert feedback for our visualization framework.
\end{abstract}


\section{INTRODUCTION}
\label{sec:intro}  
Colonoscopy is used to detect polyps, ulcers, and inflammation in the colon. Under anesthesia, tube is introduced through the rectum and the colon is traversed to the cecum (insertion phase) and back (withdrawal phase), using a video relayed to an external monitor from a tiny camera attached to the tip of this tube. Traditionally, the video of the procedure is not captured and stored and hence, an important piece of prognosis (follow-up procedure) information is lost. This was due to storage limitations on the picture archive and communication system (PACS) and the lack of visualization tools to analyze these videos. Recently, with the advances in databases, electronic medical record (EMR) regulations, and lawsuits against hospitals for missed polyps, the storage of endoscopy videos has become a high priority for many hospitals. The challenge now is to develop effective visualization tools to analyze these videos, to generate post-procedure reports (traditionally done via recalling the information on part of the endoscopist and hence, the possibility of missing out on minute yet essential details due to recall bias) and to compare against the videos from the follow-up procedures for the same patient and against a specific patient population.  We present such a tool in this work and present important use cases and medical expert feedback for this tool.

More than 14 million colonoscopies are performed each year in the USA and the majority of these videos are not captured or stored for later analysis \cite{seeff:2004}. These videos can be crucial in follow-up procedures, which happen frequently in the case of Inflammatory Bowel Diseases (IBDs) in the population ranging from age 2--21 \cite{kirschner:1988}, as well as in the case of colorectal cancer screening which is advised by the American Cancer Society to be done every 5 years for patients of age 50 and older \cite{winawer:2003}. In the case of colorectal cancer, even though the techniques have improved, the rate of cancer remains high because of polyp miss rates as well as incomplete removal of polyps due to which the cancer comes back (interval cancer). A visualization tool to facilitate a tight bound localization of an anomaly from an earlier procedure can minimize anesthetic time and unproductive colon examination and hence improve overall patient care.

Due to recent lawsuits for missed polyps \cite{barclay:2006} and the new regulations demanding comprehensive reporting of the procedure for the patient's EMR, hospitals are trying to improve the quality of the endoscopy procedures and have recently started capturing and storing the endoscopy videos. In this context, a computer-aided detection (CAD) module can help confirm the diagnosis by automatically identifying the polyps in the endoscopy video and later, ease the process of documentation for the patient's EMR. An objective quality metric, such as withdrawal time (which has been correlated with polyp miss rate), if reported, can also help in improving the overall quality of the endoscopy procedure.

Report generation/documentation for these procedures is another problematic avenue and can take up a significant amount of the physician's time. The reporting phase is subject to recall bias whereby a physician may leave out details (both small or large) when a post-procedure note is written after the colonoscopy is finished, especially when videos are not captured and stored for verification of details. However, even when the videos are stored this can be a tedious task. If the important colonoscopy features can be annotated and the findings and impressions correspondingly tagged while going through the video, the report can be generated in a semi-automated fashion. The physician can later verify the report and fill in the missing details.

\setlength{\tabcolsep}{1.2pt}
\begin{figure}[ht!]
\begin{center}
\begin{tabular}{cc}
\includegraphics[width=0.35\textwidth]{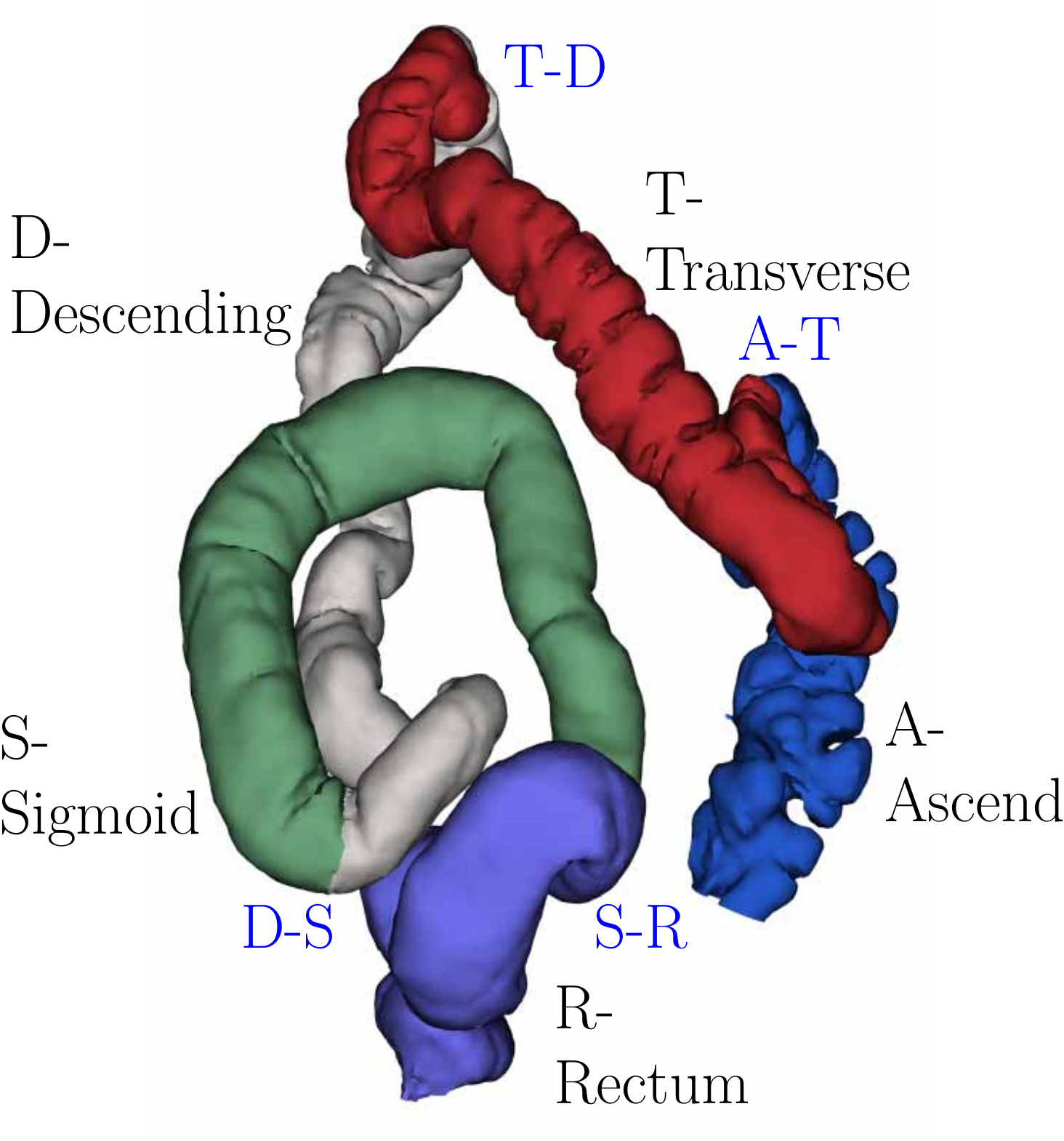}&
\includegraphics[width=0.35\textwidth]{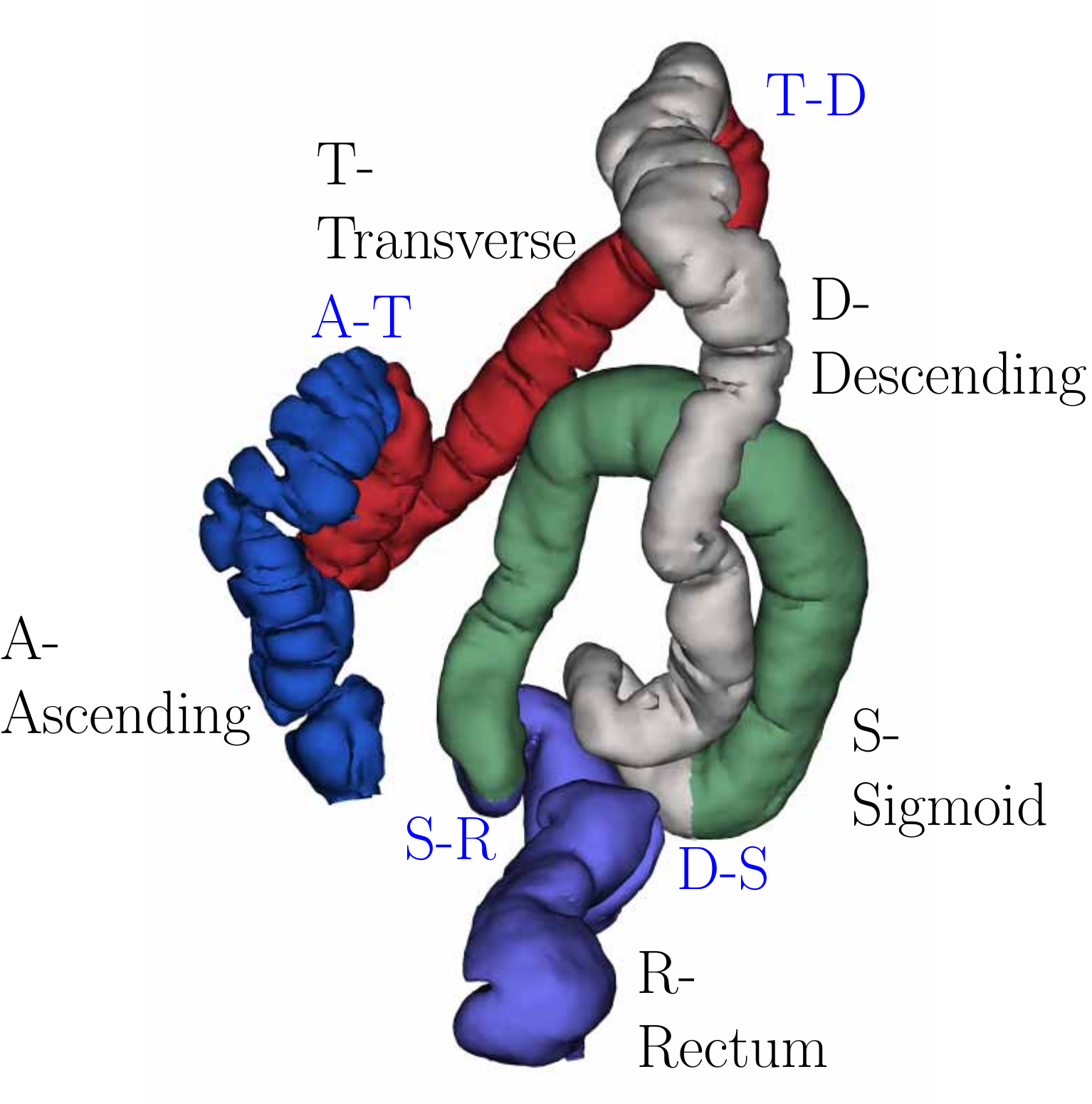}
\end{tabular}
\end{center}
\caption{The front and back views of four flexures on a generic colon model, which divide the colon surface into five color-coded segments.}
\label{fig:flexures}
\end{figure}

\section{VISUALIZATION FRAMEWORK}
Our framework uses audio synced with the colonoscopy videos to help add tags and annotations to the procedure. The framework is implemented using the Java Swing library. The setting for the endoscopy suite is as follows: There are four people in the endoscopy suite (endoscopist, supervising doctor, anesthesiologist, and a nurse) - five in the case of a teaching hospital (with a medical student being the fifth). The endoscopist has to verbally say-out-loud (1) all his findings (lesions, etc.) to the supervising doctor in the room who will confirm the diagnosis based on the findings, and (2) report distances-from-anus, based on the markings (in 5cm granularity) on the endoscope.

\setlength{\tabcolsep}{2pt}
\begin{figure}[h]
\begin{center}
\begin{tabular}{cccc}
\includegraphics[width=0.23\textwidth,height=0.23\textwidth]{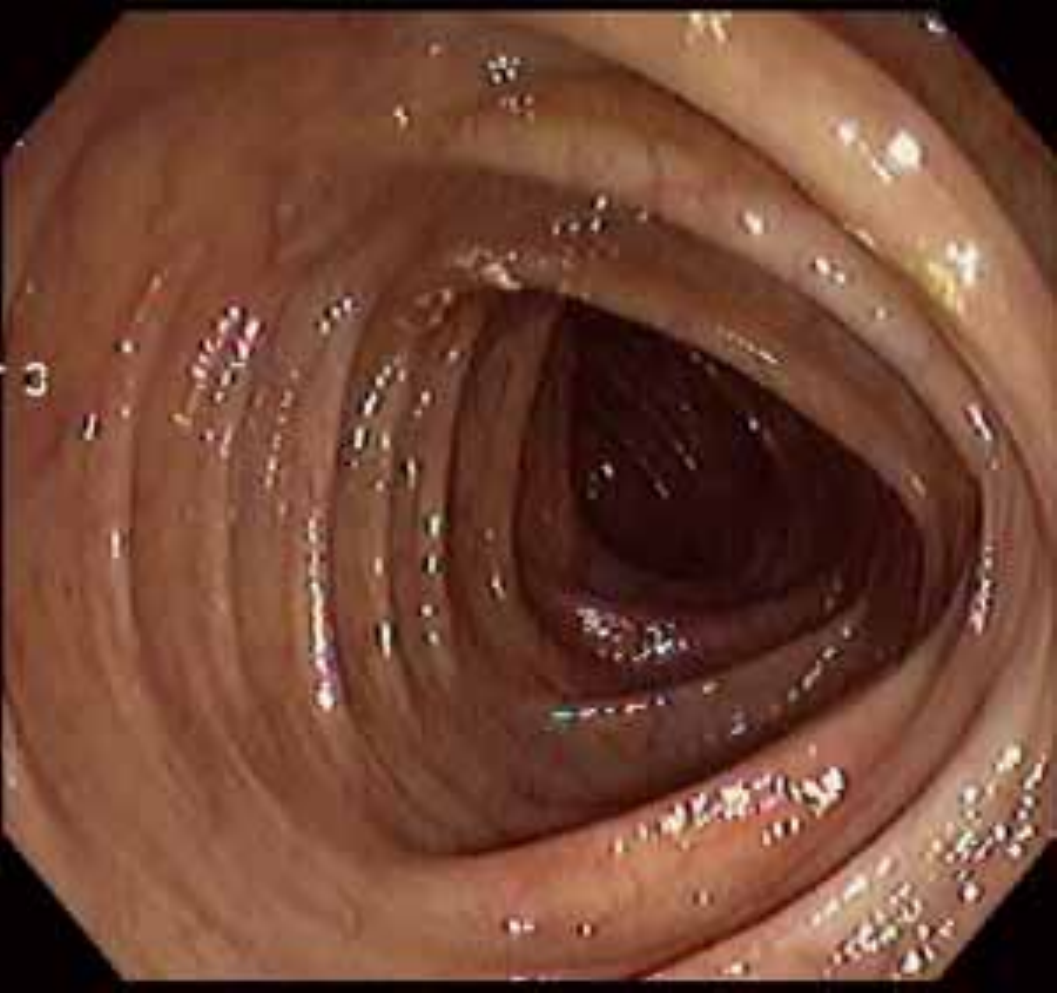}&
\includegraphics[width=0.23\textwidth,height=0.23\textwidth]{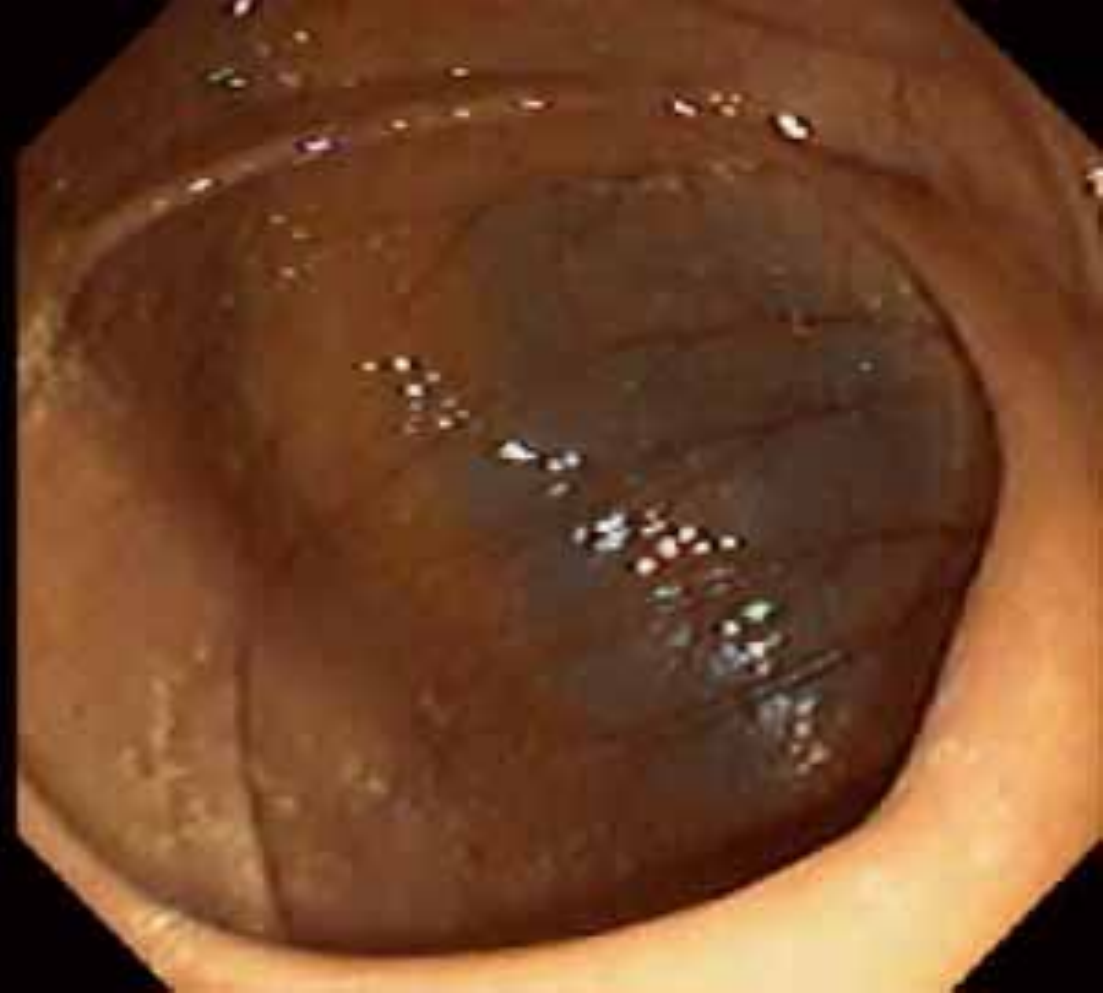}&
\includegraphics[width=0.23\textwidth,height=0.23\textwidth]{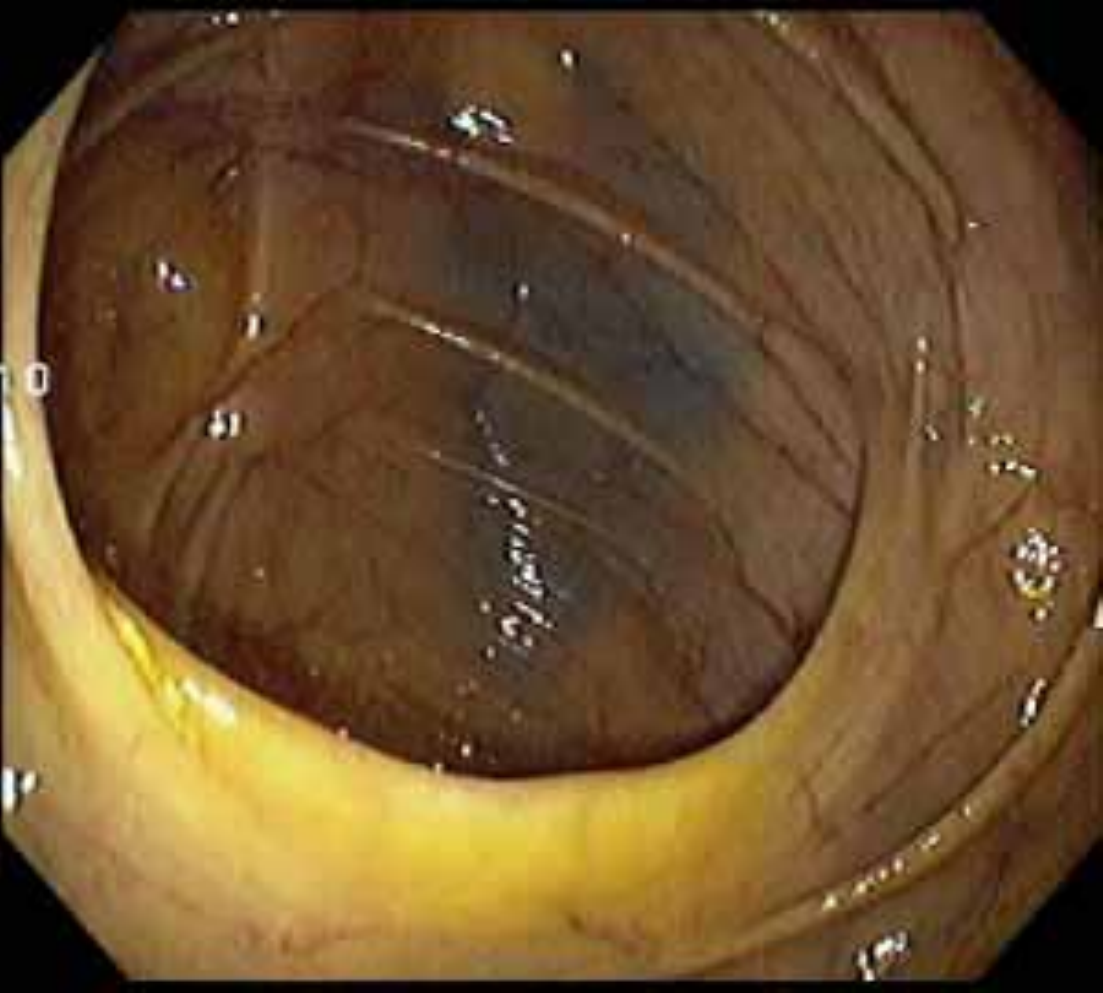}&
\includegraphics[width=0.23\textwidth,height=0.23\textwidth]{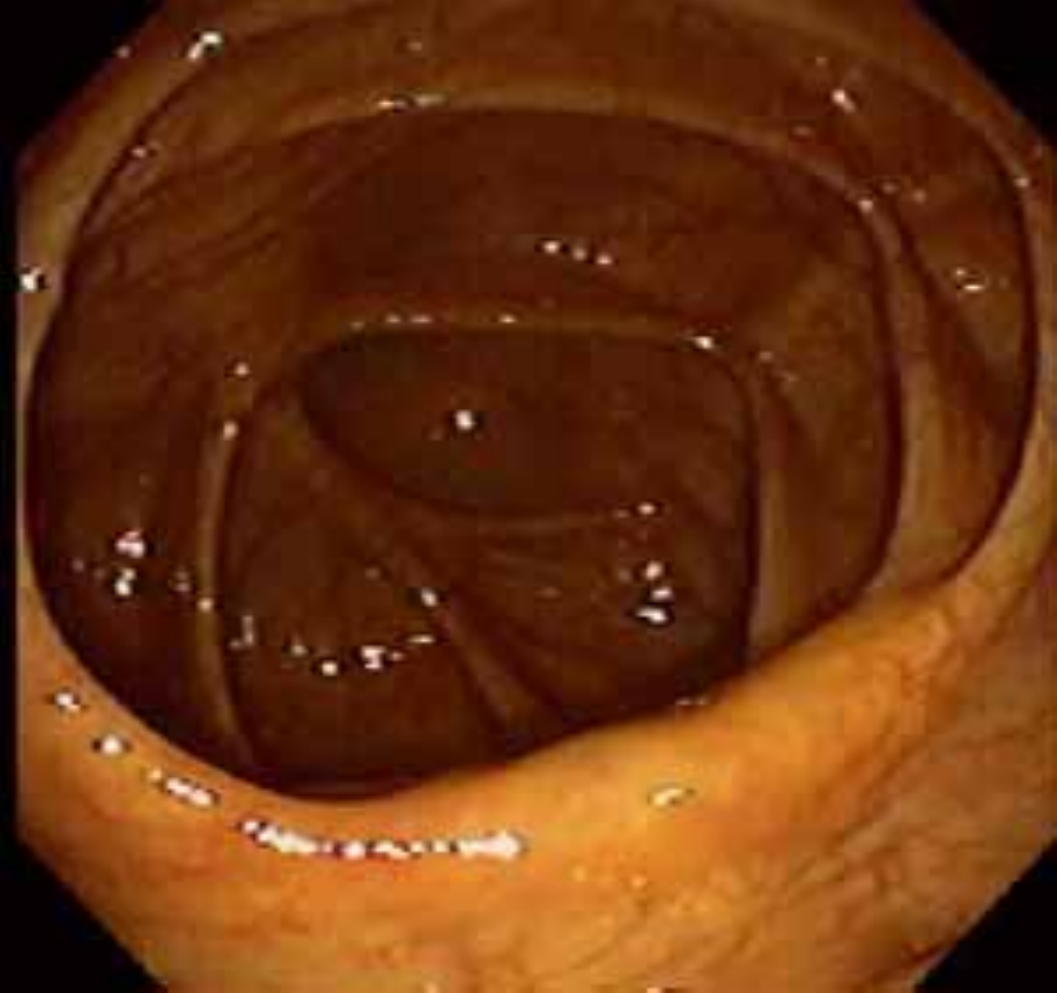}\\
(a) & (b) & (c) & (d)\\
\end{tabular}
\end{center}
\caption{Colonoscopy video frames showing important landmarks for annotation. (a) Transverse colon has the characteristic of triangular fold contour patterns. (b) Splenic flexures can be identified with the bluish discoloration on the colon wall, indicating proximity to spleen. (c) Hepatic flexures can be identified with bluish discoloration on the colon wall, indicating proximity to liver. (d) Cecum can be identified using the ileocecal valve juncture and the appendiceal orifice (small opening on the right to the appendix).}
\label{fig:endoscope}
\end{figure}

\subsection{Annotation and Tagging}
\begin{figure}[ht!]
\begin{center}
\begin{tabular}{cc}
\includegraphics[height=0.6\textwidth]{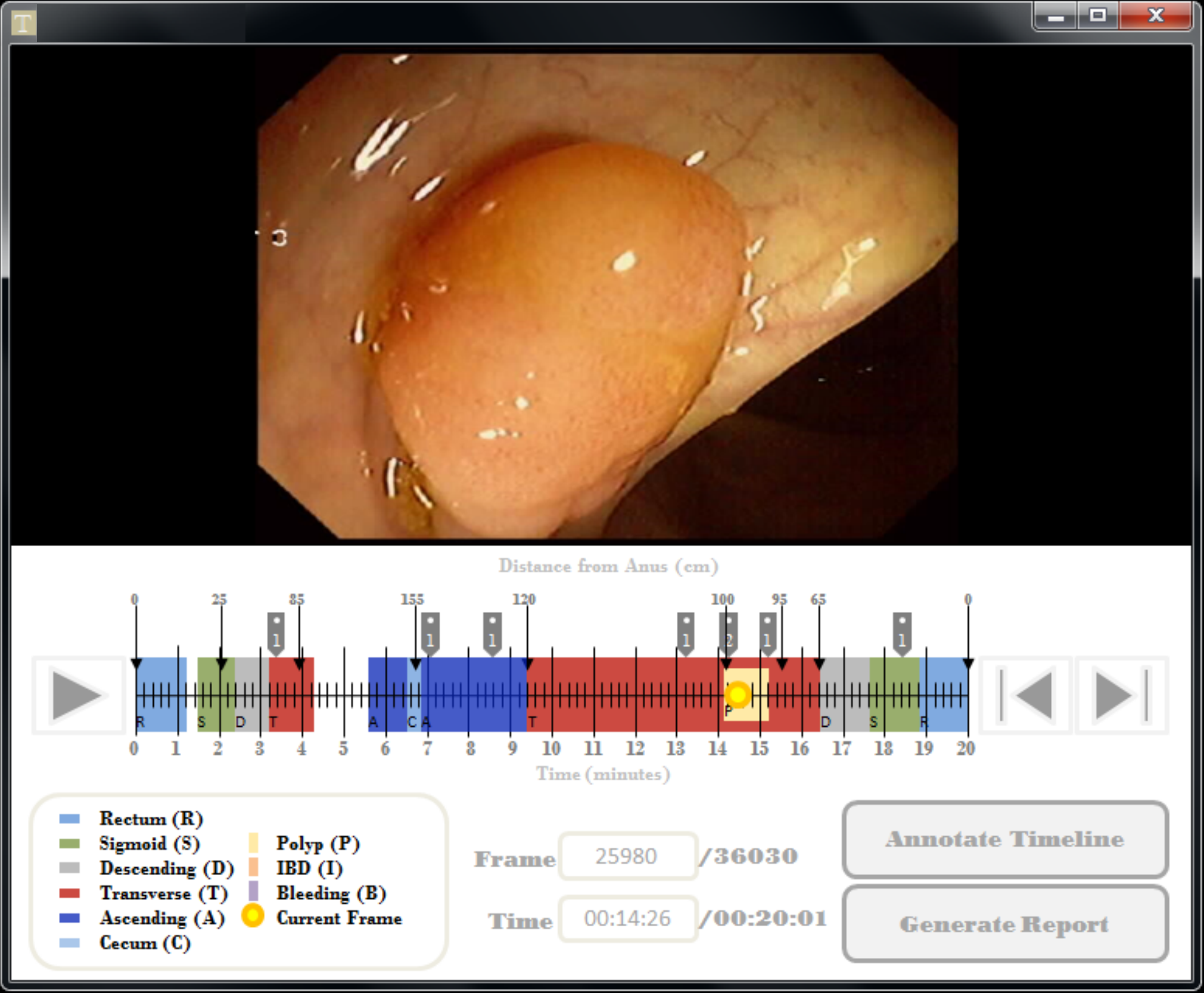} &
\includegraphics[height=0.6\textwidth]{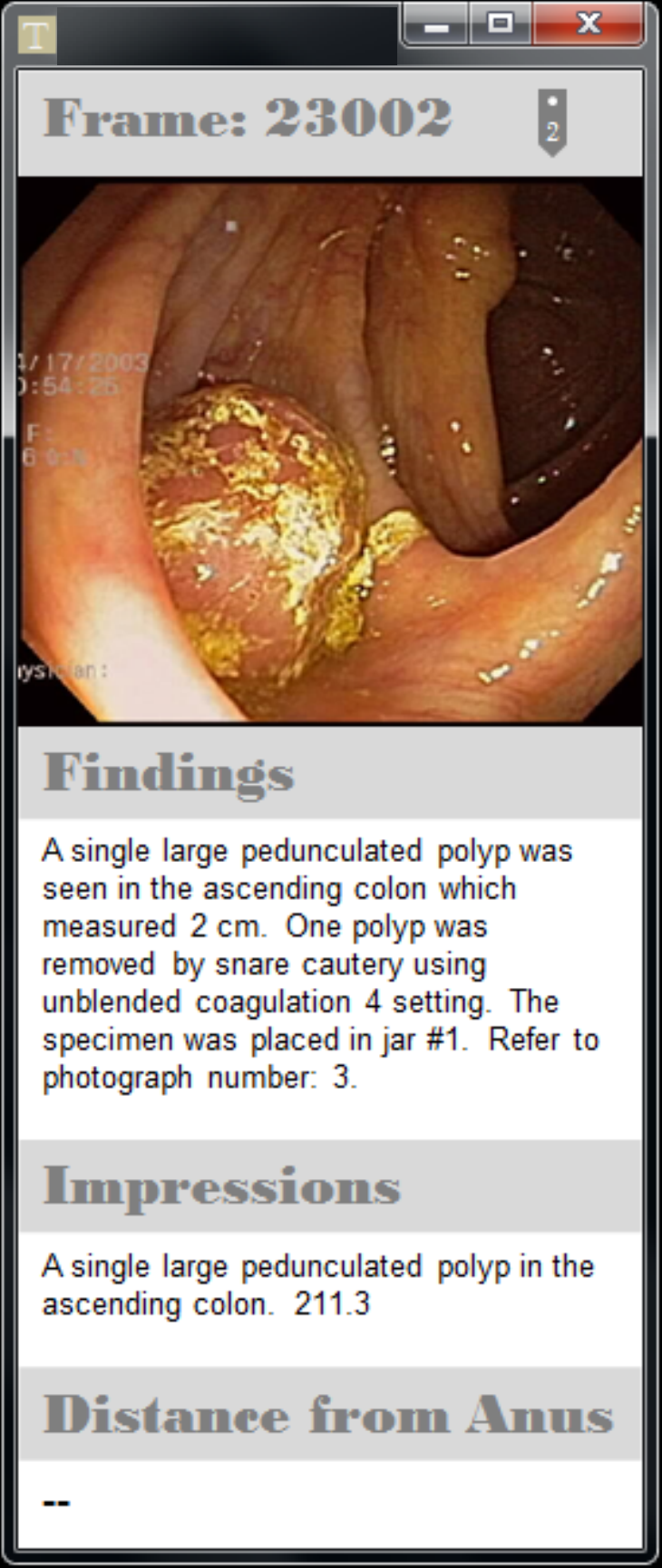} \\
\multicolumn{2}{c}{\includegraphics[width=1\textwidth]{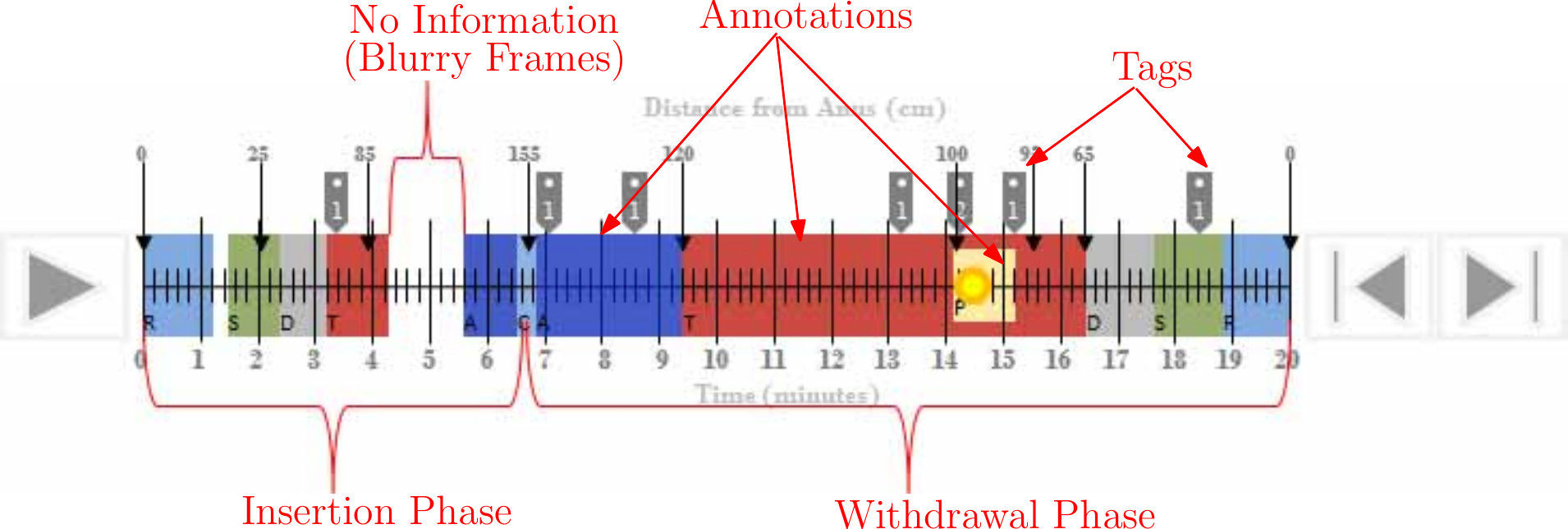}}\\
\end{tabular}
\end{center}
\caption{Visualization interface for annotating and tagging colonoscopy videos. Given an input video, the endoscopist or an assistant can tag the distances from the synchronized audio and annotate colon regions (R, S, D, T, A, C), as identified in Figure \ref{fig:flexures}, and anomalies (polyps (P), IBDs (I) or bleeding (B)) during the insertion and withdrawal phase. The tagged distances are in 5cm granularity because this is the accuracy at which the endoscopist can read on the endoscope markings. The endoscopist can also add impressions and findings corresponding to different tags. Certain regions in these videos are sometimes hard to identify because of blurry frames, as can be seen in the above timeline.
\label{fig:interface}}
\end{figure}

Given an input colonoscopy video and synced audio (identifying distances-from-anus, segments, and anomalies), we present the user with the timeline view, as shown in Figure \ref{fig:interface}, for annotation and tagging of other additional elements shown in the bottom left key. The user can add annotations on the timeline by clicking on the ``annotate timeline'' button at the bottom right of Figure \ref{fig:interface}, clicking on the timeline at a certain instance (indicating the start of the annotation) and then dragging to the appropriate end frame for this annotation. Once the cursor is released at the end frame, a drop down menu pops up at the cursor allowing the user to select one of the annotations, rectum (R), sigmoid (S), descending (D), transverse (T), ascending (A), cecum (C), polyp (P), IBD (I), or blood clot (B). Similarly, a user can add the tags for distances, impressions and findings to a specific frame (using the synced audio) by simply right clicking on that frame and selecting the tag option. This pops open a separate window (the second window on the right in Figure \ref{fig:interface}) which allows the user to add findings, impressions and distance-from-anus. If the user only adds the distance and leaves the findings and impression fields blank, the distance mark is added to the timeline (without the tag). However, if one or both the findings and impressions are added, the tag is added to the timeline. The distances-from-anus are in multiples of 5 since the endoscope markings are at 5cm granularity.

When the user annotates a set of frames, they are color-coded according to the key given at the bottom left in Figure \ref{fig:interface}. Moreover the annotations are added in hierarchical fashion to the timeline with colon segments occupying the outermost hierarchy, followed by polyps, IDBs and bleeding. This hierarchy keeps the timeline visualization easier to follow and to compare across other timelines, as shown in Figure \ref{fig:comp}.

Even though some portions of the video can be blurry due to fluid motion or camera movement, the user can still mark the regions based on the landmarks identified (see Figure \ref{fig:endoscope}). The splenic flexure is a sharp bend between the transverse and descending colon segments and exhibit strong blue discoloration pressing on the colon wall (in colonoscopy images), indicating the proximity of the spleen. This can be used to break the timeline between transverse (T) and descending (D) segments. Similarly, hepatic flexures is a sharp bend between the ascending and descending colon segments and exhibit strong blue discoloration pressing on the colon wall (in colonoscopy images), indicating the proximity of the liver. Hence, this can likewise be used to break the timeline between ascending (A) and descending (D) colon segments. Furthermore, different colon segments have different fold contour characteristics, for example, the transverse colon segment has triangular fold contours as compared to more circular fold contours in other colon segments. The rectum and cecum are the two extreme points of the large intestine, also called the colon. Hence the user can mark the rectum at the beginning of the insertion phase and at the end of the withdrawal phase on the video timeline. The cecum, on the other hand, can be marked where the ileocecal valve (sphincter muscle valve that separates the large and small intestines) and the appendiceal orifice (small opening that connects the appendix to the cecum) landmarks are identified. These two landmarks indicate that a insertion phase has been completed and the withdrawal phase can start. We use the cecum annotation to automatically compute the total insertion and withdrawal times for a given colonoscopy procedure, as shown in Figure \ref{fig:comp}.

Currently, the user has to annotate the important features manually. In the future, we will work on methods to automatically detect polyps and therapeutic tools in the colonoscopy frames, like the ones presented in Wang et al. \cite{wang:2014}, Nawarathna et al. \cite{nawarathna:2014} and Tajbakhsh et al. \cite{tajbaksh:2015}. This will expedite the process of annotation and report generation.

\subsection{Documentation of Endoscopy Videos}
\begin{figure}[ht!]
\begin{center}
\begin{tabular}{ccc}
\includegraphics[height=0.45\textwidth]{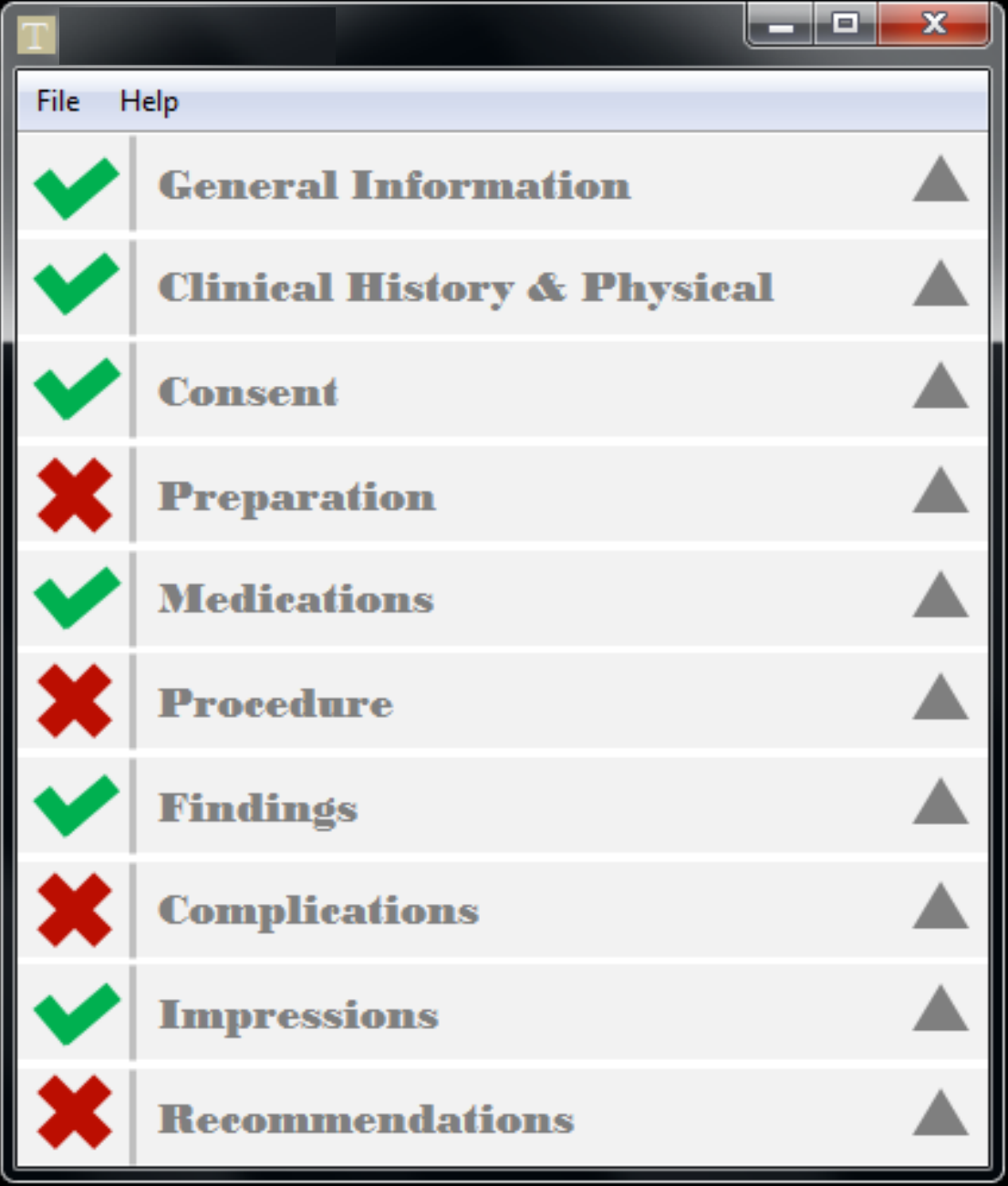}&
\includegraphics[height=0.45\textwidth]{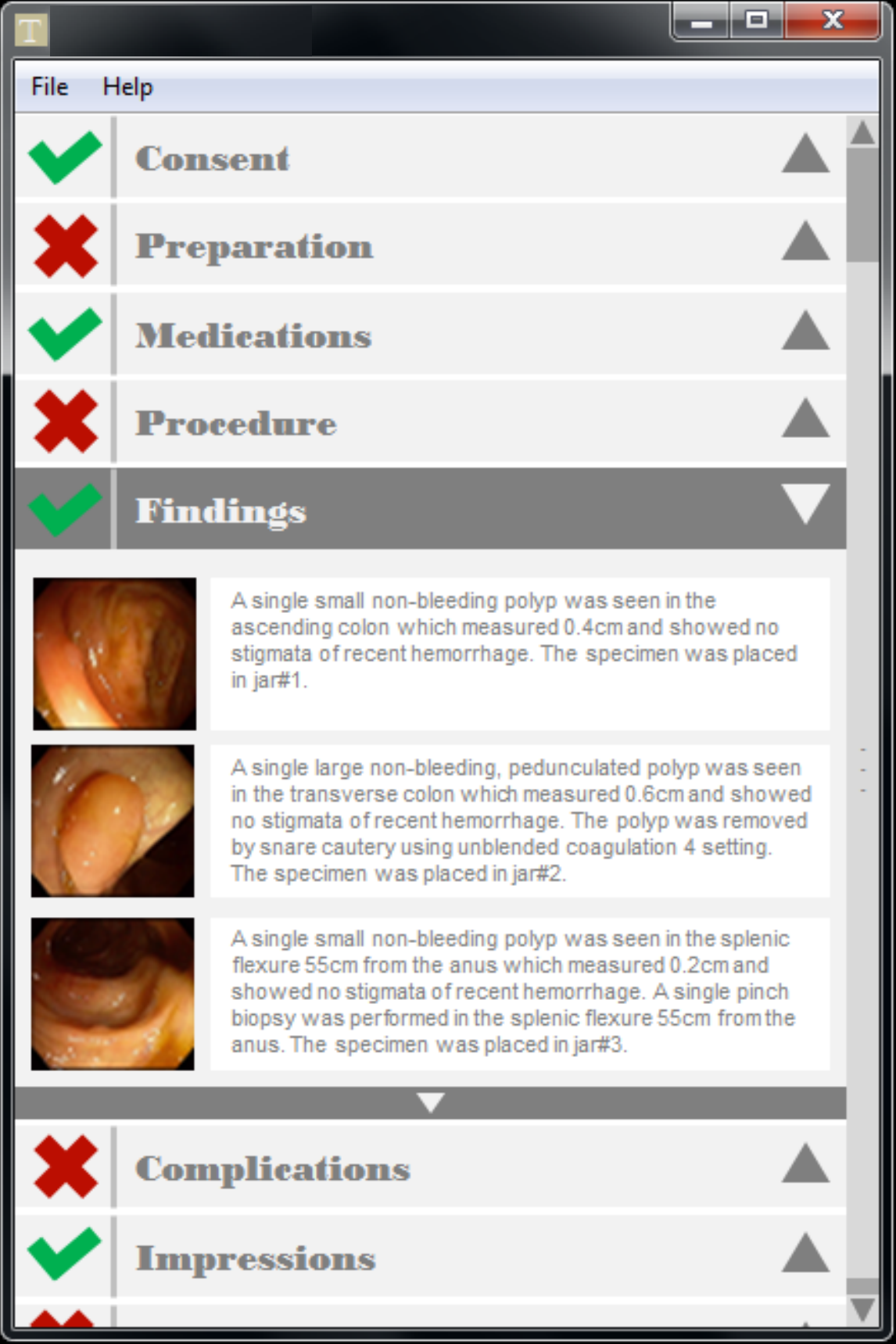}&
\includegraphics[height=0.45\textwidth]{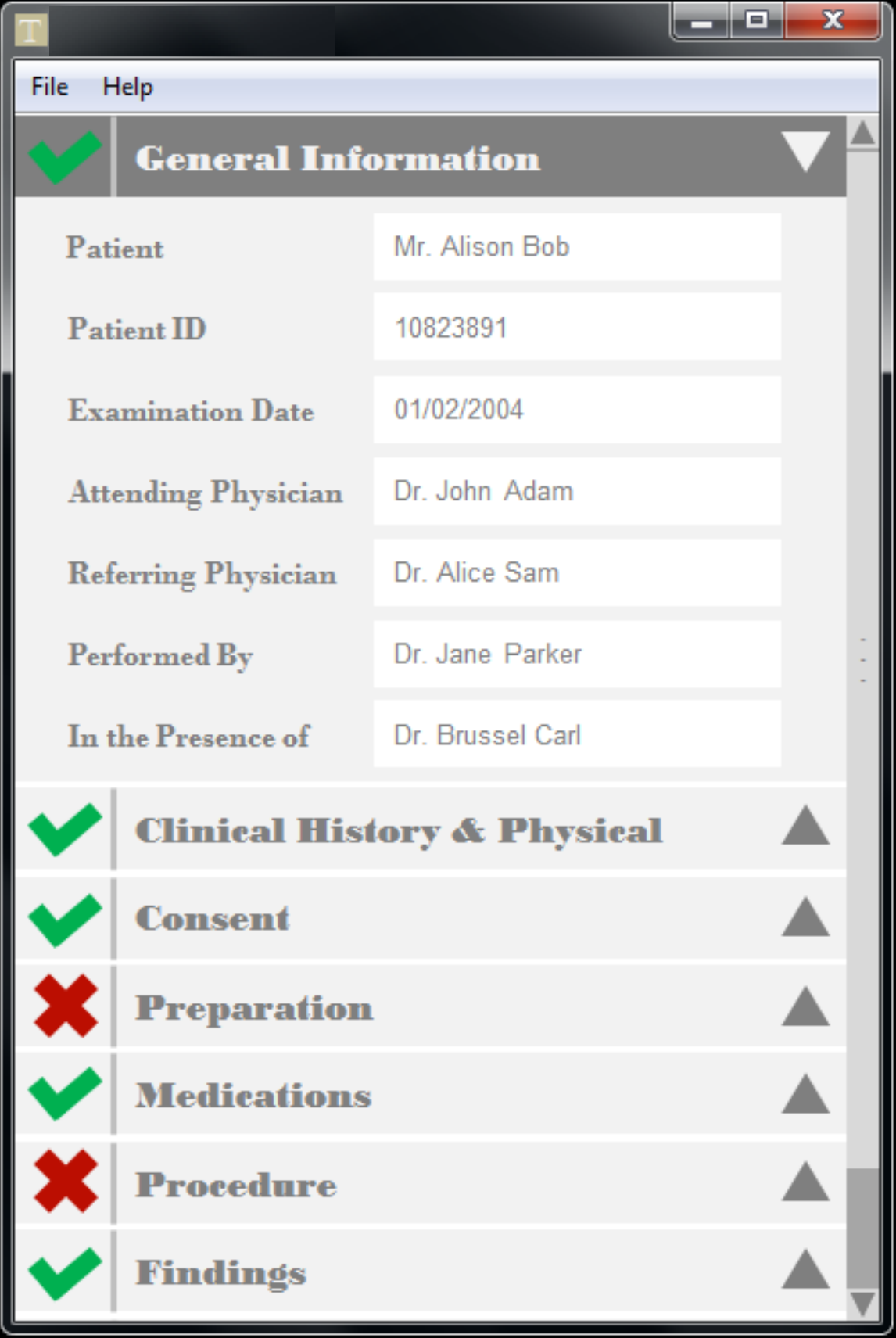}\\
\end{tabular}
\end{center}
\caption{Documentation Tool. Multiple views of the same window are shown with different drop-down expanded views. General information, clinical history and physicals, consent, medications, findings and impressions can be generated from the tags, annotations and the prior patient information. Preparation, procedure, and complications during the procedure, and recommendations are added separately.}
\label{fig:documentation}
\end{figure}

Once the physician has annotated the colon segments (rectum, sigmoid, descending, transverse, ascending, and cecum), anomalies (polyps, IBDs, bleeding), tagged findings and distances, these annotations and tags can be used for semi-automatic report generation for the patient's EMR, as shown in Figure \ref{fig:documentation} by clicking the ``generate report'' button, shown in Figure \ref{fig:interface}. For example, the polyp location with respect to a segment, its respective tagged distance, findings and impressions, and an initial frame snapshot can be used to fill in most of the fields in the report without the recall bias, and with high accuracy. The final recommendation for the patient, any complications incurred during the procedure, and the initial patient preparation can be added at the end. All these details are then verified by the physician and once verified, marked complete and added to the patient's EMR.

\setlength{\tabcolsep}{1.2pt}

\subsection{Comparison of Endoscopy Videos}
\label{sec:comp}
\begin{figure}[ht!]
\begin{center}
\includegraphics[width=1\textwidth]{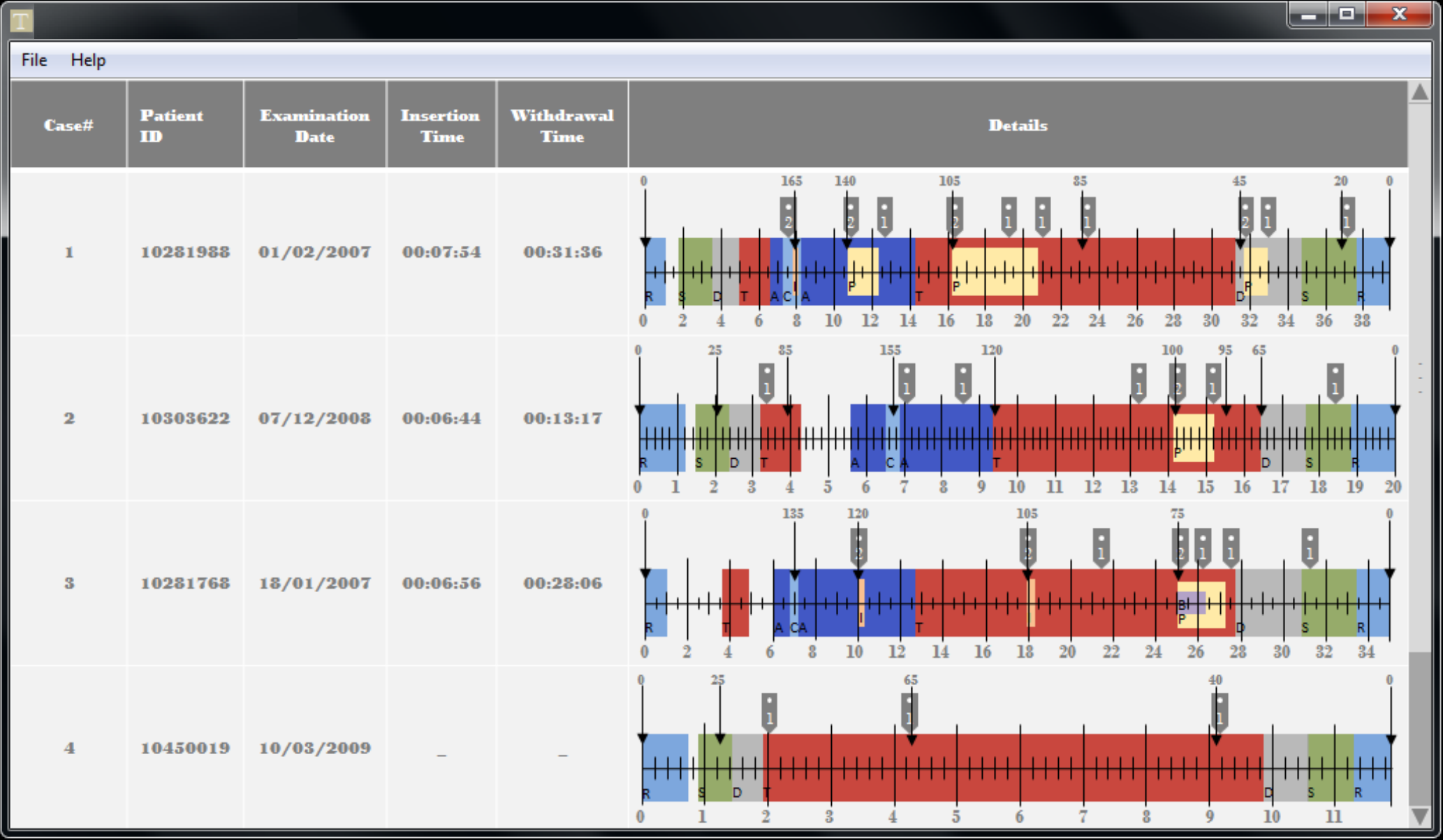}
\end{center}
\caption{Comparison Module: Different cases with annotations and tags can be compared. Once a tag is clicked, the video window with the tagged frame will open such as the one shown on the right in Figure \ref{fig:interface}. The details of these cases are explained in Section \ref{sec:comp}.}
\label{fig:comp}
\end{figure}

The hierarchical annotations and tags can be used to compare different procedures at an individual as well as at a population level. A sample of cases in the comparison module is shown in Figure \ref{fig:comp}. Case \#1 details show a completed colonoscopy procedure with 1 IBD found in the cecum segment (165cm from anus) and 3 polyps found in the ascending (140cm from anus), transverse (105cm from anus), and descending colon segments (45cm from anus). Case \#2 details show a completed colonoscopy procedure with 1 polyp found in the transverse colon segment (100cm from anus). Case \#3 details show a completed colonoscopy procedure with 2 IBDs found in the ascending (120cm from anus) and the transverse colon segments (105cm from anus) and 1 polyp with blood clot found in the transverse colon segment (75cm from anus). Case \#4 details show an incomplete colonoscopy procedure which was followed by a surgery to examine the cause of obstruction. Since the cecum was not annotated, the insertion and withdrawal times were not computed. Moreover, as observed from the table in Figure \ref{fig:comp} the insertion time is less than the withdrawal time, since the protocol that is normally followed is that the endoscope is quickly navigated to the cecum and then careful examination is done in the withdrawal phase. The careful examination in the withdrawal phase also allows the endoscopist to identify different flexures and colon segments based on the color and the fold contour characteristics, as shown in Figure \ref{fig:endoscope}. The missing information in the insertion phase in these cases is attributed to the blurry frames mainly because of the quick camera motion. Fluid motion can be a cause of blurry flames as well but it was not the case in these four colonoscopy procedures. We have also created timeline visualizations for a few colonoscopy videos from the publicly available National Institute of Biomedical Imaging and Bioengineering (NIBIB) Image and Clinical Data Repository provided by the National Institute of Health (NIH) using the available detailed colonoscopy examination reports containing findings, impressions, and the corresponding distances-from-anus and the snapshots of the colonoscopy video frames. The timeline visualizations can help in analyzing the data from the thousands of colonoscopy videos available in the repository at a population as well at an individual level.

\section{Medical Expert Feedback}

We interviewed gastroenterologists for feedback on our system. These included one gastroenterologist who performs colonoscopies on adults (of age 50 years and older) and one from the pediatric gastroenterology department, who performs colonoscopies on patients in the 2--21 year old age group to screen for IBDs such as Crohn's disease. Some specific anecdotes and comments are as follows:
\begin{itemize}
    \item This technology would definitely benefit gastroenterologists. A typical colonoscopy lasts anywhere from thirty minutes to an hour and a half. Because of this, small but ultimately important details may be left out of the post-procedure report. For example, we recently performed a follow-up colonoscopy on a 15 year-old male with Crohn's Disease because we wanted to see how the patient's condition had changed after a year on medication. The original report stated that inflammation was in the cecum, ascending, and transverse colon. During the repeat endoscopy, we once again found inflammation in the cecum and ascending colon but when we got to the transverse colon half of it was healthy mucosa. We were unable to determine how much of the colon was inflamed at the time of the original report. Perhaps if we had such an annotation software, during the initial procedure we could have noted that the inflammation ended at, for example, the second distal haustral fold.
    \item Since the endoscopist has to speak out loud his/her findings to the supervisor or gastroenterologist present in the room, maybe a speech-to-text program can provide the annotation and tagging information required in our framework, automatically, during the procedure. This can save a significant amount of time in the post-procedure analysis and documentation. It might also make the gastroenterologists more vigilant when doing the procedure since they will have to focus on even finer details.
\end{itemize}

\section{CONCLUSION}
We have presented a visualization framework to annotate and tag colonoscopy videos. These annotations and tags can help analyze colonoscopy videos, generate reports, and compare patient populations. In the future, we will plan to automate the tagging process from the audio by using automatic speech recognition based on the voice tags. We are also in the process of acquiring a ScopeGuide, an HD endoscope fitted with electromagnetic sensors, in collaboration with the Department of Pediatric Gastroenterology at Stony Brook University Children's Hospital, to provide millimeter level accuracy for distance tags in real time during the procedure.

\section{ACKNOWLEDGEMENTS}
We would like to thank Dr James Brief of the Department of Pediatric Gastroenterology, Stony Brook University Children's Hospital, and Dr Jeffrey Morganstern of Department of Gastroenterology, Stony Brook Hospital, for their help and guidance with this project.



\begin{thebibliography}{1}

\bibitem{seeff:2004}
Seeff, L.~C., Richards, T.~B., Shapiro, J.~A., Nadel, M.~R., Manninen, D.~L.,
  Given, L.~S., Dong, F.~B., Winges, L.~D., and McKenna, M.~T., ``How many
  endoscopies are performed for colorectal cancer screening? {R}esults from
  {CDC�s} survey of endoscopic capacity,'' {\em Gastroenterology}~{\bf
  127}(6),  1670--1677 (2004).

\bibitem{kirschner:1988}
Kirschner, B., ``Inflammatory bowel disease in children.,'' {\em Pediatric
  Clinics of North America}~{\bf 35}(1),  189--208 (1988).

\bibitem{winawer:2003}
Winawer, S., Fletcher, R., Rex, D., Bond, J., Burt, R., Ferrucci, J., Ganiats,
  T., Levin, T., Woolf, S., Johnson, D., Kirk, L., Litin, S., and Simmang, C.,
  ``Colorectal cancer screening and surveillance: clinical guidelines and
  rationale update based on new evidence,'' {\em Gastroenterology}~{\bf
  124}(2),  544--560 (2003).

\bibitem{barclay:2006}
Barclay, R.~L., Vicari, J.~J., Doughty, A.~S., Johanson, J.~F., and Greenlaw,
  R.~L., ``Colonoscopic withdrawal times and adenoma detection during screening
  colonoscopy,'' {\em New England Journal of Medicine}~{\bf 355}(24),
  2533--2541 (2006).

\bibitem{wang:2014}
Wang, Y., Tavanapong, W., Wong, J., Oh, J., and De~Groen, P.~C., ``Part-based
  multiderivative edge cross-sectional profiles for polyp detection in
  colonoscopy,'' {\em IEEE Journal of Biomedical and Health Informatics}~{\bf
  18}(4),  1379--1389 (2014).

\bibitem{nawarathna:2014}
Nawarathna, R., Oh, J., Muthukudage, J., Tavanapong, W., Wong, J., De~Groen,
  P.~C., and Tang, S.~J., ``Abnormal image detection in endoscopy videos using
  a filter bank and local binary patterns,'' {\em Neurocomputing}~{\bf 144},
  70--91 (2014).

\bibitem{tajbaksh:2015}
Tajbakhsh, N., Gurudu, S., and Liang, J., ``Automated polyp detection in
  colonoscopy videos using shape and context information,'' {\em IEEE
  Transactions on Medical Imaging}~{\bf PP}(99),  1--16 (2015).

\end{thebibliography}
\end{document}